\pdfoutput=1

\documentclass[11pt]{article}

\usepackage[]{emnlp2021}

\usepackage{times}
\usepackage{latexsym}
\usepackage{graphicx}

\usepackage[T1]{fontenc}

\usepackage[utf8]{inputenc}

\usepackage{microtype}
\usepackage{booktabs}

\title{Tiny Neural Models for Seq2Seq}

\author{Arun Kandoor \\
  Google Research \\
  Mountain View, CA 94043 \\
  \texttt{akandoor@google.com}}

\begin{document}
\maketitle
\begin{abstract}

Semantic parsing models with applications in task oriented dialog systems require efficient sequence to sequence (seq2seq) architectures to be run on-device.  To this end, we propose a projection based encoder-decoder model referred to as pQRNN-MAtt. Studies based on projection methods were restricted to encoder-only models, and we believe this is the first study extending it to seq2seq architectures. The resulting quantized models are less than 3.5MB in size and are well suited for on-device latency critical applications. We show that on MTOP  \citep{DBLP:journals/corr/abs-2008-09335}, a challenging multilingual semantic parsing dataset, the average model performance surpasses LSTM based seq2seq model that uses pre-trained embeddings despite being 85x smaller. Furthermore, the model can be an effective student \citep{DBLP:journals/corr/abs-2101-08890} for distilling large pre-trained models such as T5/BERT \citep{DBLP:journals/corr/abs-1910-10683}.

\end{abstract}

\section{Introduction}
Privacy concerns and connectivity issues have spurred interest in on-device neural applications. Neural semantic parsing is one such problem that converts natural language into machine executable logical forms usable in applications such as voice assistant. Though there is much research on advancing the state of the art in neural semantic parsing, there is little research on achieving the same high quality results within the compute and memory capabilities of edge devices.

Neural seq2seq models employ an encoder-decoder model, in which the encoder converts word tokens to a latent representation. This is then fed to a decoder to generate output tokens based on a target vocabulary. Initial seq2seq models employed an architecture where all the input sequence information is encoded into one single state and is provided to the decoder for generating target sequence \citep{DBLP:journals/corr/SutskeverVL14}. More recent approaches such as \citet{bahdanau2016neural} employed attention mechanism, which makes use of  $\mathit{all}$ the encoder outputs and thereby improving the model performance.

Current state of the art models for semantic parsing are attention based pointer networks \citep{45283,DBLP:journals/corr/abs-2001-11458,DBLP:journals/corr/abs-2008-09335} and achieve impressive performance on server with minimal to no restrictions on the model size and inference times. The same models however, would not be suitable for inference on edge devices. 

Another aspect of these models is the use of embedding tables for word representations. Embedding tables increase in size as vocabulary increases and would not scale well for on-device applicaitons. As previous studies have demonstrated \citep{kaliamoorthi-etal-2019-prado,DBLP:journals/corr/abs-2101-08890,DBLP:journals/corr/abs-1708-00630,ravi-kozareva-2018-self-governing}, text projection is an effective alternative to embedding tables for on-device problems in natural language processing. In this work, we extend text projections to seq2seq problems such as neural semantic parsing by combining them with efficient decoder architectures.

A main motivation in this work is to identify effective neural encoding and decoding architectures that operate on textual input and are suitable for on-device applications. In our experiments and based on previous work \citep{DBLP:journals/corr/abs-2101-08890}, projections used with QRNN \citep{DBLP:journals/corr/BradburyMXS16} encoder proves to be a effective combination for text classification and labeling tasks. We extend this model to include a Merged Attention (MAtt) decoder \citep{zhang-etal-2019-improving-deep} and demonstrate that the resulting model architecture we refer to as pQRNN-MAtt is a promising candidate for on-device neural semantic parsing and code generation tasks. Experiments on multilingual MTOP dataset show that average exact match accuracy for pQRNN-MAtt model is higher than LSTM models with pre-trained XLU embeddings, despite the former being 85x smaller than latter.

\section{Related Work}

Recent work on neural semantic parsing is largely based on encoder-decoder models and has shown promising results on tasks such as machine translation \citep{DBLP:journals/corr/SutskeverVL14} and image captioning \citep{DBLP:journals/corr/VinyalsTBE14, bahdanau2016neural}. \citet{DBLP:journals/corr/LuongPM15} improved these architectures by implementing an attention mechanism in the decoder.

One major drawback with these models was its inability to learn good enough parameters for long tail entities. This was addressed with the advent of Pointer Networks \cite{45283}, in which the decoder decides to either copy a token from the input query or generate a token from the output vocabulary. \citet{DBLP:journals/corr/abs-2001-11458,DBLP:journals/corr/abs-2008-09335} employed this model to achieve impressive results on public datasets.

All these studies incorporate architectures based on recurrent neural networks or Transformers \citep{DBLP:journals/corr/VaswaniSPUJGKP17} and use some form of pre-trained token representations \cite{DBLP:journals/corr/abs-1712-09405}, which are not well suited for on-device applications as the model size is dominated by the embedding table. Large embedding table is necessary to reach high quality with these model architectures.

Projection based methods \citep{kaliamoorthi-etal-2019-prado,DBLP:journals/corr/abs-2101-08890,DBLP:journals/corr/abs-1708-00630,ravi-kozareva-2018-self-governing} have been studied extensively for on-device applications essentially replacing embedding tables with hashing based techniques. While promising results have been shown on problems that can be solved with a neural encoder, no one has studied the applicability of these methods for seq2seq tasks like semantic parsing.

To address this shortcoming, we complement the projection based methods with efficient decoder architectures like Merged Attention (MAtt) \citep{zhang-etal-2019-improving-deep} and study the overall performance of the model for semantic parsing task.

\section{Model Architecture}

\begin{figure}
\centering
\includegraphics[width=0.45\textwidth]{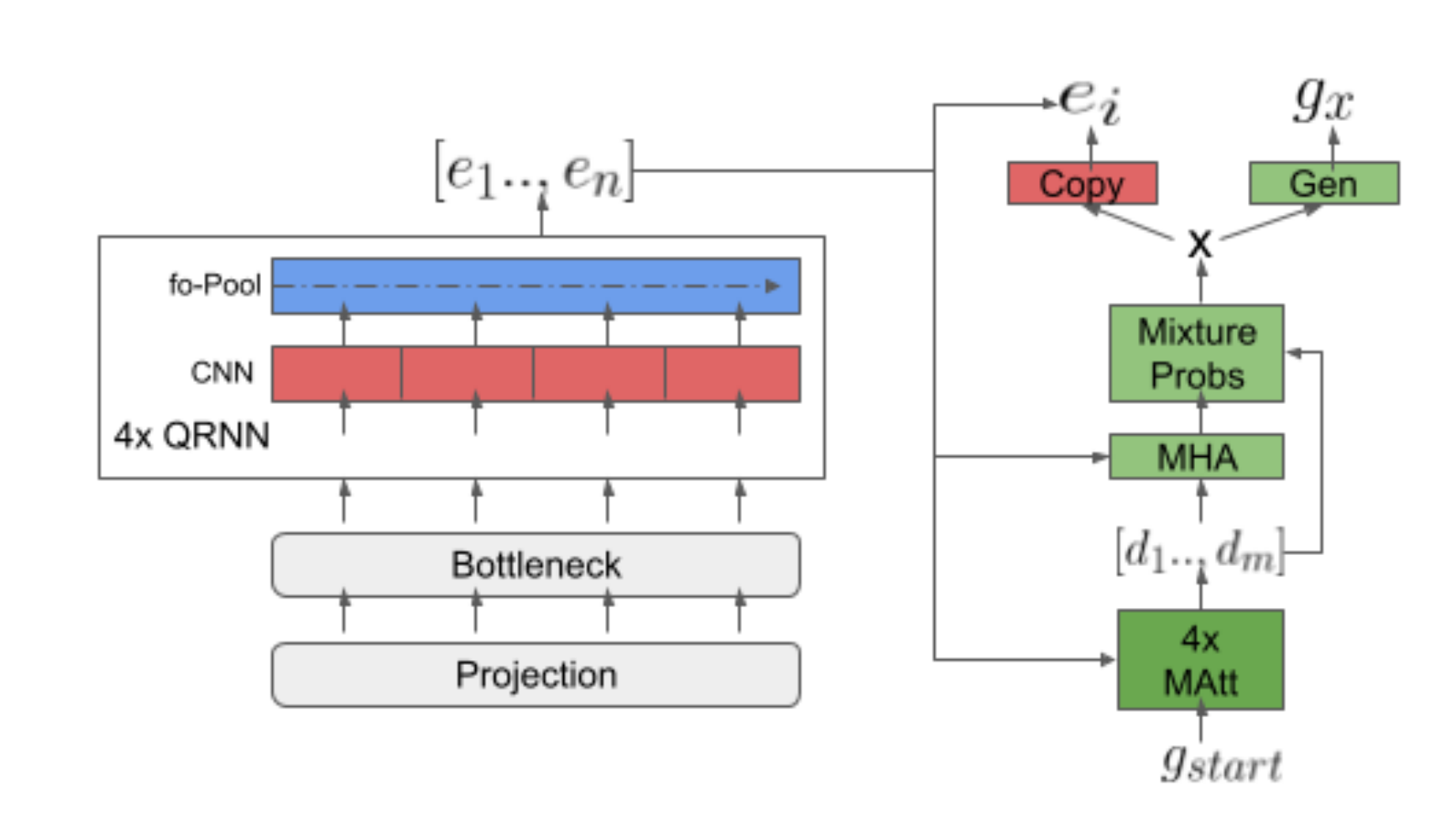}
\caption{
Pointer generator network with pQRNN encoder and MAtt decoder.}
\label{fig:enc_dec}
\end{figure}

Our model architecture combines pQRNN encoder \citet{DBLP:journals/corr/abs-2101-08890} with MAtt decoder \citet{zhang-etal-2019-improving-deep} and design it as a pointer-generator network as proposed in \citet{DBLP:journals/corr/abs-2001-11458}.

As illustrated in Figure \ref{fig:enc_dec}, the encoder block consists of projection stage that converts source tokens to a sequence of ternary vectors \citet{DBLP:journals/corr/abs-2101-08890}. The ternary representation is then fed to a dense layer (bottleneck) with $\mathit{ReLU}$ activation. Since the projection features are not trainable, the bottleneck layer allows the network to learn semantic similarity needed for the task. A stack of bidirectional QRNN \citep{DBLP:journals/corr/BradburyMXS16} is then used to learn a contextual representation for the input.

An important modification to the decoder is, since projections lack semantic and contextual information, the encoder hidden states $[e_1, e_2, ...,e_n]$ are used as decoder input embeddings for copy tokens. So while decoding at time step $t$, if the previous decoded token is a copy token at encoder step $i$, the corresponding encoder hidden state $e_i$ is chosen to be the input to the decoder. 

On the decoder output, at step $t$ and with hidden states
$[d_1..,d_m]$ , as proposed in \citet{DBLP:journals/corr/abs-2008-09335}, the final output distribution is a mixture of generation and copy distributions:
\[p_t = p_t^w . p_t^g + (1-p_t^w) * p_t^c \] where \[p_t^w = \mbox{\emph{sigmoid}}(Linear_\alpha[d_t;w_t])\] $p_t^g$, $p_t^c$ and $w_t$ are computed as: \[p_t^g=\mbox{\emph{softmax}}(Linear_g[d_t])\] \[p_t^c, w_t = \mathit{MultiHeadAttn}(e_1...,e_n;Linear_c[d_t])\] 

Figure \ref{fig:enc_dec} illustrates the end-to-end flow for decoding at step $0$. The embedding for decoder output token $x$, which is the input for next step, is decided based on whether $x$ is copy or generate token.

\begin{table*}
\centering
\resizebox{0.85\textwidth}{!}{
\begin{tabular}{l|c|cccccc|c}
\toprule
            &           & \multicolumn{6}{c|}{Exact Match Accuracy} &       \\
pQRNN+MAtt  & \#Params          & en   & es   & fr   & de   & hi   & th   & avg  \\
\midrule
Top1        &   3.3M (8bit)     & 77.8 & 69.9 & 66.6 & 63.7 & 64.5 & 64.3   & 67.8  \\
Top2        &   3.3M (8bit)     & 81.2 & 73.7 & 70.9 & 68.7 & 68.3 & 69.2   & 72.0 \\
Top3        &   3.3M (8bit)     & 81.9 & 74.8 & 72.0 & 70.1 & 69.6 & 70.5   & 73.2 \\
Top4        &   3.3M (8bit)     & 82.2 & 75.3 & 72.3 & 70.6 & 70.3 & 70.9   & 73.6 \\

\midrule
                \multicolumn{9}{c}{Reference} \\
\midrule
XLU         & 70M (float)       & 77.8 & 66.5 & 65.6 & 61.5 & 61.5 & 62.8   & 66.0 \\
XLM-R       & 550M (float)      & 83.9 & 76.9 & 74.7 & 71.2 & 70.2 & 71.2   & 74.7 \\
\bottomrule
\end{tabular}
}
\caption{Exact Match Accuracy on compositional decoupled representation for 6 languages. Reference numbers have been taken from Table 3 \cite{DBLP:journals/corr/abs-2008-09335}.}
\label{tab:exactmatch}
\end{table*}

\begin{table*}
\centering
\resizebox{0.85\textwidth}{!}{
\begin{tabular}{l|cccccc}
\toprule
                            & \multicolumn{6}{c}{Intent Accuracy / Slot F1}      \\
pQRNN+MAtt                  & en        & es        & fr        & de        & hi        & th       \\
                \midrule
Top1                        & 94.7/88.0 & 93.1/80.1 & 91.1/78.4 & 91.5/76.2 & 91.0/76.5 & 91.5/79.5     \\
Top2                        & 95.6/89.2 & 94.0/81.6 & 92.5/80.0 & 92.7/78.6 & 92.0/78.3 & 92.7/81.3     \\
Top3                        & 95.9/89.4 & 94.3/82.0 & 92.9/80.6 & 93.2/79.2 & 92.5/78.7 & 93.1/81.9     \\
Top3                        & 96.0/89.5 & 94.4/82.2 & 93.0/80.7 & 93.5/79.4 & 92.7/79.1 & 93.2/82.1     \\
\bottomrule
\end{tabular}
}
\caption{ Intent Accuracy / Slot F1 for models in Table \ref{tab:exactmatch}.}
\label{tab:intent}
\end{table*}

\subsection{Quantization}

Effective quantization techniques allow end-to-end models to run inference using integer-only arithmetic and reduce model footprints. We adapted the quantization scheme proposed in \citet{DBLP:journals/corr/abs-1712-05877}, which allows us to simulate quantization during training and learn the ranges for weights and activations in the model.

In this setup, training happens in floating point arithmetic, but the forward pass also simulates 8-bit integer quantization using fake-quantized tensorflow ops \citep{tensorflow2015-whitepaper} which allows us to collect weights and activation stats. Later, Tensorflow Lite converter tool uses these stats to construct an 8-bit Tensorflow Lite model, which is used for running inference.

\section{Experiments}

We evaluate the model performance using MTOP dataset from \citet{DBLP:journals/corr/abs-2008-09335} on all 6 languages using only target language training data. Exact match accuracy, intent accuracy and slot F1 metrics are reported for all the models. As we could not verify whether the metrics presented in \citet{DBLP:journals/corr/abs-2008-09335} were Top1 result from the decoder output, we chose to present TopK (K=4) results for comparison. We conduct the experiments using the compositional decoupled representaion as labels.

\subsection{Model configuration}
The model uses open source projection operator \footnote{\url{https://github.com/tensorflow/models/tree/master/research/seq\_flow\_lite}}
with feature dimension $N=1024$. The projection output is then fed to a dense layer (bottleneck) with output width $256$. The dense layer output is then fed to the QRNN stack of $4$ layers, each with state size $128$ and convolution kernel width set to $2$.

The decoder input embedding size is set to $128$, which is followed by MAtt decoder stack of size $4$. Each decoder has model dimension set to $128$ and the hidden dimension in the feed forward network is set to $1024$. We averaged across 4 heads when computing copy probabilities using $\mathit{MultiHeadAttn}$.

\section{Results}

Table \ref{tab:exactmatch} shows the TopK metrics on exact match accuracy for all 6 languages on the compositional decoupled representation. On average, Top1 results outperform the LSTM baseline model with XLU embeddings from \citet{DBLP:journals/corr/abs-2008-09335}. For K>1, the exact match accuracy approaches close to the large pre-trained model XLM-R.

The model effectiveness is indicated by the Params column, which roughly can be mapped to the model footprint and inference times.

Table \ref{tab:intent} shows the top level intent accuracy and slot F1 metrics for TopK results for compositional decoupled representation. \citet{DBLP:journals/corr/abs-2008-09335} doesn't 
consider this combination, but overall the model proves to be effective in generating the intent and arguments pertaining to the given query.

\section{Conclusion}
We extend Projection based representations to on-device seq2seq models using QRNN encoder and MAtt decoder. Despite being 85x smaller, evaluations on MTOP dataset proved the model to be highly effective when compared to LSTM models trained with pre-trained embeddings.

Future directions include employing distillation techniques  \citep{DBLP:journals/corr/abs-2101-08890} to improve the model further and exploring different tokenization schemes for multilingual projections.

\section*{Acknowledgements}

We would like to thank our colleagues Prabhu Kaliamoorthi, Erik Vee, Edgar Gonzàlez i Pellicer, Evgeny Livshits, Ashwini Venkatesh, Derik Clive, Edward Li, Milan Lee and the Learn2Compress team for
helpful discussions related to this work. We would also like to thank Amarnag Subramanya, Andrew Tomkins and Rushin Shah for their leadership and support.

\bibliography{pqrnn_matt}

\begin{thebibliography}{18}
\expandafter\ifx\csname natexlab\endcsname\relax\def\natexlab#1{#1}\fi

\bibitem[{Abadi et~al.(2015)Abadi, Agarwal, Barham, Brevdo, Chen, Citro,
  Corrado, Davis, Dean, Devin, Ghemawat, Goodfellow, Harp, Irving, Isard, Jia,
  Jozefowicz, Kaiser, Kudlur, Levenberg, Man\'{e}, Monga, Moore, Murray, Olah,
  Schuster, Shlens, Steiner, Sutskever, Talwar, Tucker, Vanhoucke, Vasudevan,
  Vi\'{e}gas, Vinyals, Warden, Wattenberg, Wicke, Yu, and
  Zheng}]{tensorflow2015-whitepaper}
Mart\'{\i}n Abadi, Ashish Agarwal, Paul Barham, Eugene Brevdo, Zhifeng Chen,
  Craig Citro, Greg~S. Corrado, Andy Davis, Jeffrey Dean, Matthieu Devin,
  Sanjay Ghemawat, Ian Goodfellow, Andrew Harp, Geoffrey Irving, Michael Isard,
  Yangqing Jia, Rafal Jozefowicz, Lukasz Kaiser, Manjunath Kudlur, Josh
  Levenberg, Dandelion Man\'{e}, Rajat Monga, Sherry Moore, Derek Murray, Chris
  Olah, Mike Schuster, Jonathon Shlens, Benoit Steiner, Ilya Sutskever, Kunal
  Talwar, Paul Tucker, Vincent Vanhoucke, Vijay Vasudevan, Fernanda Vi\'{e}gas,
  Oriol Vinyals, Pete Warden, Martin Wattenberg, Martin Wicke, Yuan Yu, and
  Xiaoqiang Zheng. 2015.
\newblock \href {https://www.tensorflow.org/} {{TensorFlow}: Large-scale
  machine learning on heterogeneous systems}.
\newblock Software available from tensorflow.org.

\bibitem[{Bahdanau et~al.(2016)Bahdanau, Cho, and Bengio}]{bahdanau2016neural}
Dzmitry Bahdanau, Kyunghyun Cho, and Yoshua Bengio. 2016.
\newblock \href {http://arxiv.org/abs/1409.0473} {Neural machine translation by
  jointly learning to align and translate}.

\bibitem[{Bradbury et~al.(2016)Bradbury, Merity, Xiong, and
  Socher}]{DBLP:journals/corr/BradburyMXS16}
James Bradbury, Stephen Merity, Caiming Xiong, and Richard Socher. 2016.
\newblock \href {http://arxiv.org/abs/1611.01576} {Quasi-recurrent neural
  networks}.
\newblock \emph{CoRR}, abs/1611.01576.

\bibitem[{Jacob et~al.(2017)Jacob, Kligys, Chen, Zhu, Tang, Howard, Adam, and
  Kalenichenko}]{DBLP:journals/corr/abs-1712-05877}
Benoit Jacob, Skirmantas Kligys, Bo~Chen, Menglong Zhu, Matthew Tang, Andrew~G.
  Howard, Hartwig Adam, and Dmitry Kalenichenko. 2017.
\newblock \href {http://arxiv.org/abs/1712.05877} {Quantization and training of
  neural networks for efficient integer-arithmetic-only inference}.
\newblock \emph{CoRR}, abs/1712.05877.

\bibitem[{Kaliamoorthi et~al.(2019)Kaliamoorthi, Ravi, and
  Kozareva}]{kaliamoorthi-etal-2019-prado}
Prabhu Kaliamoorthi, Sujith Ravi, and Zornitsa Kozareva. 2019.
\newblock \href {https://doi.org/10.18653/v1/D19-1506} {{PRADO}: Projection
  attention networks for document classification on-device}.
\newblock In \emph{Proceedings of the 2019 Conference on Empirical Methods in
  Natural Language Processing and the 9th International Joint Conference on
  Natural Language Processing (EMNLP-IJCNLP)}, pages 5012--5021, Hong Kong,
  China. Association for Computational Linguistics.

\bibitem[{Kaliamoorthi et~al.(2021)Kaliamoorthi, Siddhant, Li, and
  Johnson}]{DBLP:journals/corr/abs-2101-08890}
Prabhu Kaliamoorthi, Aditya Siddhant, Edward Li, and Melvin Johnson. 2021.
\newblock \href {http://arxiv.org/abs/2101.08890} {Distilling large language
  models into tiny and effective students using pqrnn}.
\newblock \emph{CoRR}, abs/2101.08890.

\bibitem[{Li et~al.(2020)Li, Arora, Chen, Gupta, Gupta, and
  Mehdad}]{DBLP:journals/corr/abs-2008-09335}
Haoran Li, Abhinav Arora, Shuohui Chen, Anchit Gupta, Sonal Gupta, and Yashar
  Mehdad. 2020.
\newblock \href {http://arxiv.org/abs/2008.09335} {{MTOP:} {A} comprehensive
  multilingual task-oriented semantic parsing benchmark}.
\newblock \emph{CoRR}, abs/2008.09335.

\bibitem[{Luong et~al.(2015)Luong, Pham, and
  Manning}]{DBLP:journals/corr/LuongPM15}
Minh{-}Thang Luong, Hieu Pham, and Christopher~D. Manning. 2015.
\newblock \href {http://arxiv.org/abs/1508.04025} {Effective approaches to
  attention-based neural machine translation}.
\newblock \emph{CoRR}, abs/1508.04025.

\bibitem[{Mikolov et~al.(2017)Mikolov, Grave, Bojanowski, Puhrsch, and
  Joulin}]{DBLP:journals/corr/abs-1712-09405}
Tom{\'{a}}s Mikolov, Edouard Grave, Piotr Bojanowski, Christian Puhrsch, and
  Armand Joulin. 2017.
\newblock \href {http://arxiv.org/abs/1712.09405} {Advances in pre-training
  distributed word representations}.
\newblock \emph{CoRR}, abs/1712.09405.

\bibitem[{Raffel et~al.(2019)Raffel, Shazeer, Roberts, Lee, Narang, Matena,
  Zhou, Li, and Liu}]{DBLP:journals/corr/abs-1910-10683}
Colin Raffel, Noam Shazeer, Adam Roberts, Katherine Lee, Sharan Narang, Michael
  Matena, Yanqi Zhou, Wei Li, and Peter~J. Liu. 2019.
\newblock \href {http://arxiv.org/abs/1910.10683} {Exploring the limits of
  transfer learning with a unified text-to-text transformer}.
\newblock \emph{CoRR}, abs/1910.10683.

\bibitem[{Ravi(2017)}]{DBLP:journals/corr/abs-1708-00630}
Sujith Ravi. 2017.
\newblock \href {http://arxiv.org/abs/1708.00630} {Projectionnet: Learning
  efficient on-device deep networks using neural projections}.
\newblock \emph{CoRR}, abs/1708.00630.

\bibitem[{Ravi and Kozareva(2018)}]{ravi-kozareva-2018-self-governing}
Sujith Ravi and Zornitsa Kozareva. 2018.
\newblock \href {https://doi.org/10.18653/v1/D18-1105} {Self-governing neural
  networks for on-device short text classification}.
\newblock In \emph{Proceedings of the 2018 Conference on Empirical Methods in
  Natural Language Processing}, pages 887--893, Brussels, Belgium. Association
  for Computational Linguistics.

\bibitem[{Rongali et~al.(2020)Rongali, Soldaini, Monti, and
  Hamza}]{DBLP:journals/corr/abs-2001-11458}
Subendhu Rongali, Luca Soldaini, Emilio Monti, and Wael Hamza. 2020.
\newblock \href {http://arxiv.org/abs/2001.11458} {Don't parse, generate! {A}
  sequence to sequence architecture for task-oriented semantic parsing}.
\newblock \emph{CoRR}, abs/2001.11458.

\bibitem[{Sutskever et~al.(2014)Sutskever, Vinyals, and
  Le}]{DBLP:journals/corr/SutskeverVL14}
Ilya Sutskever, Oriol Vinyals, and Quoc~V. Le. 2014.
\newblock \href {http://arxiv.org/abs/1409.3215} {Sequence to sequence learning
  with neural networks}.
\newblock \emph{CoRR}, abs/1409.3215.

\bibitem[{Vaswani et~al.(2017)Vaswani, Shazeer, Parmar, Uszkoreit, Jones,
  Gomez, Kaiser, and Polosukhin}]{DBLP:journals/corr/VaswaniSPUJGKP17}
Ashish Vaswani, Noam Shazeer, Niki Parmar, Jakob Uszkoreit, Llion Jones,
  Aidan~N. Gomez, Lukasz Kaiser, and Illia Polosukhin. 2017.
\newblock \href {http://arxiv.org/abs/1706.03762} {Attention is all you need}.
\newblock \emph{CoRR}, abs/1706.03762.

\bibitem[{Vinyals et~al.(2015)Vinyals, Fortunato, and Jaitly}]{45283}
Oriol Vinyals, Meire Fortunato, and Navdeep Jaitly. 2015.
\newblock \href {https://arxiv.org/pdf/1506.03134.pdf} {Pointer networks}.
\newblock In \emph{NIPS}, pages 2692--2700.

\bibitem[{Vinyals et~al.(2014)Vinyals, Toshev, Bengio, and
  Erhan}]{DBLP:journals/corr/VinyalsTBE14}
Oriol Vinyals, Alexander Toshev, Samy Bengio, and Dumitru Erhan. 2014.
\newblock \href {http://arxiv.org/abs/1411.4555} {Show and tell: {A} neural
  image caption generator}.
\newblock \emph{CoRR}, abs/1411.4555.

\bibitem[{Zhang et~al.(2019)Zhang, Titov, and
  Sennrich}]{zhang-etal-2019-improving-deep}
Biao Zhang, Ivan Titov, and Rico Sennrich. 2019.
\newblock \href {https://doi.org/10.18653/v1/D19-1083} {Improving deep
  transformer with depth-scaled initialization and merged attention}.
\newblock In \emph{Proceedings of the 2019 Conference on Empirical Methods in
  Natural Language Processing and the 9th International Joint Conference on
  Natural Language Processing (EMNLP-IJCNLP)}, pages 898--909, Hong Kong,
  China. Association for Computational Linguistics.

\end{thebibliography}
\bibliographystyle{acl_natbib}

\end{document}